\documentclass[10pt]{article} % For LaTeX2e
\usepackage[accepted]{tmlr}
\usepackage{amsmath,amsfonts,bm}

\def\eqref#1{equation~\ref{#1}}
\def\1{\bm{1}}

\DeclareMathAlphabet{\mathsfit}{\encodingdefault}{\sfdefault}{m}{sl}
\SetMathAlphabet{\mathsfit}{bold}{\encodingdefault}{\sfdefault}{bx}{n}

\usepackage{hyperref}
\usepackage{url}
\usepackage{wrapfig}

\title{Vision-and-Language Navigation Today and Tomorrow: \\A Survey in the Era of Foundation Models}

\author{%
\name Yue Zhang$^{1*}$ \email zhan1624@msu.edu
      \AND
    \name Ziqiao Ma$^{2*}$ \email 
    marstin@umich.edu
    \AND
      \name Jialu Li$^{3*}$ \email
      jialuli@cs.unc.edu
      \AND
      \name Yanyuan Qiao$^{4*}$ \email yanyuan.qiao@adelaide.edu.au
       \AND
      \name Zun Wang$^{3*}$ \email 
      zunwang@cs.unc.edu
       \AND
      \name Joyce Chai$^{2\dagger}$ \email
      chaijy@umich.edu
       \AND
      \name Qi Wu$^{4\dagger}$ \email qi.wu01@adelaide.edu.au
       \AND
      \name Mohit Bansal$^{3\dagger}$ \email 
      mbansal@cs.unc.edu
       \AND
      \name Parisa Kordjamshidi$^{1\dagger}$ \email kordjams@msu.edu 
\\
$^{1}$ Michigan State University \quad
$^{2}$ University of Michigan \quad
$^{3}$ UNC Chapel Hill \quad
$^{4}$ University of Adelaide \\
$^{*}$Equal Contribution
$^{\dagger}$ Equal Supervision
}
\def\month{12}  % Insert correct month for camera-ready version
\def\year{2024} % Insert correct year for camera-ready version
\def\openreview{\url{https://openreview.net/forum?id=yiqeh2ZYUh}} % Insert correct link to OpenReview for camera-ready version

\begin{document}

\maketitle

\begin{abstract}

Vision-and-Language Navigation (VLN) has gained increasing attention over recent years and many approaches have emerged to advance their development.
The remarkable achievements of foundation models have shaped the challenges and methods for VLN research.
In this survey, we provide a top-down review that adopts a principled framework for embodied planning and reasoning, and emphasizes the current methods and future opportunities leveraging foundation models to address VLN challenges.
We hope our in-depth discussions could provide valuable resources and insights: on the one hand, to document the progress and explore opportunities and potential roles for foundation models in this field, and on the other, to organize different challenges and solutions in VLN to foundation model researchers\footnote{GitHub repository: \url{https://github.com/zhangyuejoslin/VLN-Survey-with-Foundation-Models}}.

\end{abstract}
\section{Introduction}

Developing embodied agents that are capable of interacting with humans and their surrounding environments is one of the longstanding goals of Artificial Intelligence (AI)~\citep{nguyen2021sensorimotor,duan2022survey}. 
These AI systems hold immense potential for real-world applications to serve as multi-functional assistants in daily life, such as household robots~\citep{habitat2}, self-driving cars~\citep{hu2023_uniad}, and personal assistants~\citep{Chu2023MobileVLMA}. 
One formal problem setting to advance this research direction is Vision-and-Language Navigation~(VLN)~\citep{mattersim}, a multimodal and cooperative task that requires the agent to follow human instructions, explore 3D environments, and engage in situated communications under various forms of ambiguity. 
Over the years, VLN has been explored in both photorealistic simulators~\citep{chang2017matterport3d,habitat,xia2018gibson} and real environments~\citep{mirowski2018learning,banerjee2021robotslang}, leading to a number of benchmarks~\citep{mattersim,ku2020room,vlnce} that each presents slightly different problem formulations.

\begin{wrapfigure}{r}{0.5\textwidth}
    \centering
 \includegraphics[width=0.5\textwidth]{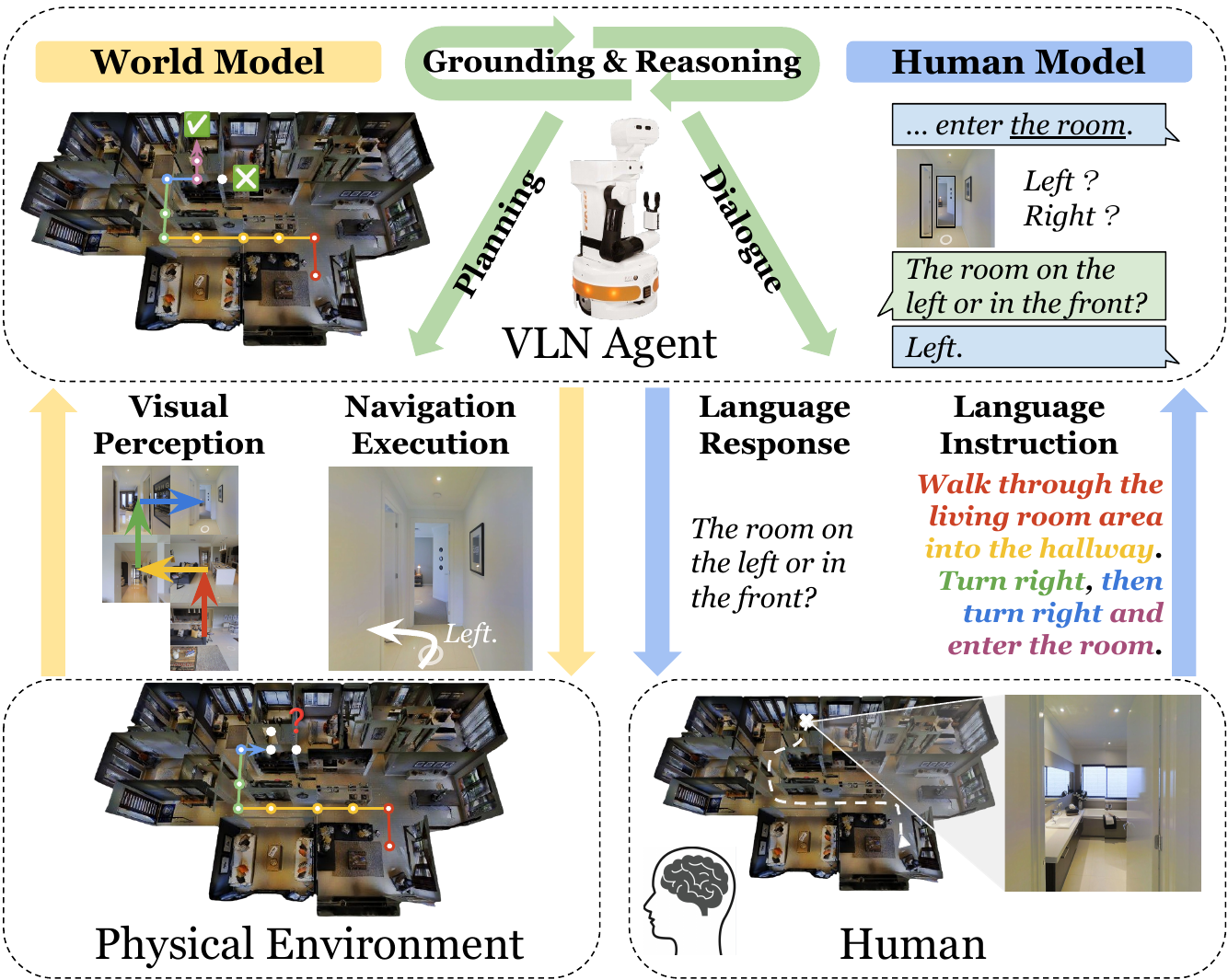}
    \caption{Organizing challenges and solutions in VLN using LAW framework~\citep{hu2023language}. 
    }
    \label{fig:world and agent}
    \vspace{-2mm}
\end{wrapfigure}

Recently, \textit{foundation models}~\citep{bommasani2021opportunities}, ranging from early pre-trained models like BERT~\citep{devlin2018bert} to contemporary large language models (LLMs) and vision-language models (VLMs)~\citep{openai2023gpt4,clip}, have exhibited exceptional abilities in multimodal comprehension, reasoning, and cross-domain generalization. 
These models are pre-trained on massive data, such as text, images, audio, and video, and could further be adapted for a broad range of specific applications, including embodied AI tasks~\citep{xu2024survey}. 
Integrating these foundation models into the VLN task marks a pivotal recent advancement for embodied AI research, demonstrated through significant performance improvements~\citep{hamt, wang2023scaling, zhou2023navgpt}.
Foundation models have also brought new opportunities to the VLN field, such as expanding the research focus from multi-modal attention learning and strategy policy learning to pre-training generic vision and language representations, hence enabling task planning, commonsense reasoning, as well as generalize to realistic environments.

Despite the recent impact of foundation models on VLN research, the previous surveys on VLN~\citep{gu2022vision,park2023visual,wu2024vision} are from the pre-foundation-model era and mainly focus on the VLN benchmarks and conventional approaches, i.e., they are missing a comprehensive overview of the existing methods and opportunities leveraging foundation models to address VLN challenges.
Especially with the emergence of LLMs, to the best of our knowledge, no review has yet discussed their applications in VLN tasks.
Moreover, unlike previous efforts that discuss the VLN task as an isolated downstream task, the objective of this survey is twofold:
\textbf{first}, to milestone the progress and explore opportunities and potential roles for foundation models in this field;
\textbf{second}, to organize different challenges and solutions in VLN to foundation model researchers within a systematic framework.
To build this connection, we adopt the LAW framework~\citep{hu2023language}, where foundation models serve as backbones of \textit{world model} and \textit{agent model}.
This framework offers a general landscape of reasoning and planning in foundation models, and is closely scoped with the core challenges in VLN.

\begin{figure*}
    \centering
    \includegraphics[width=\linewidth]{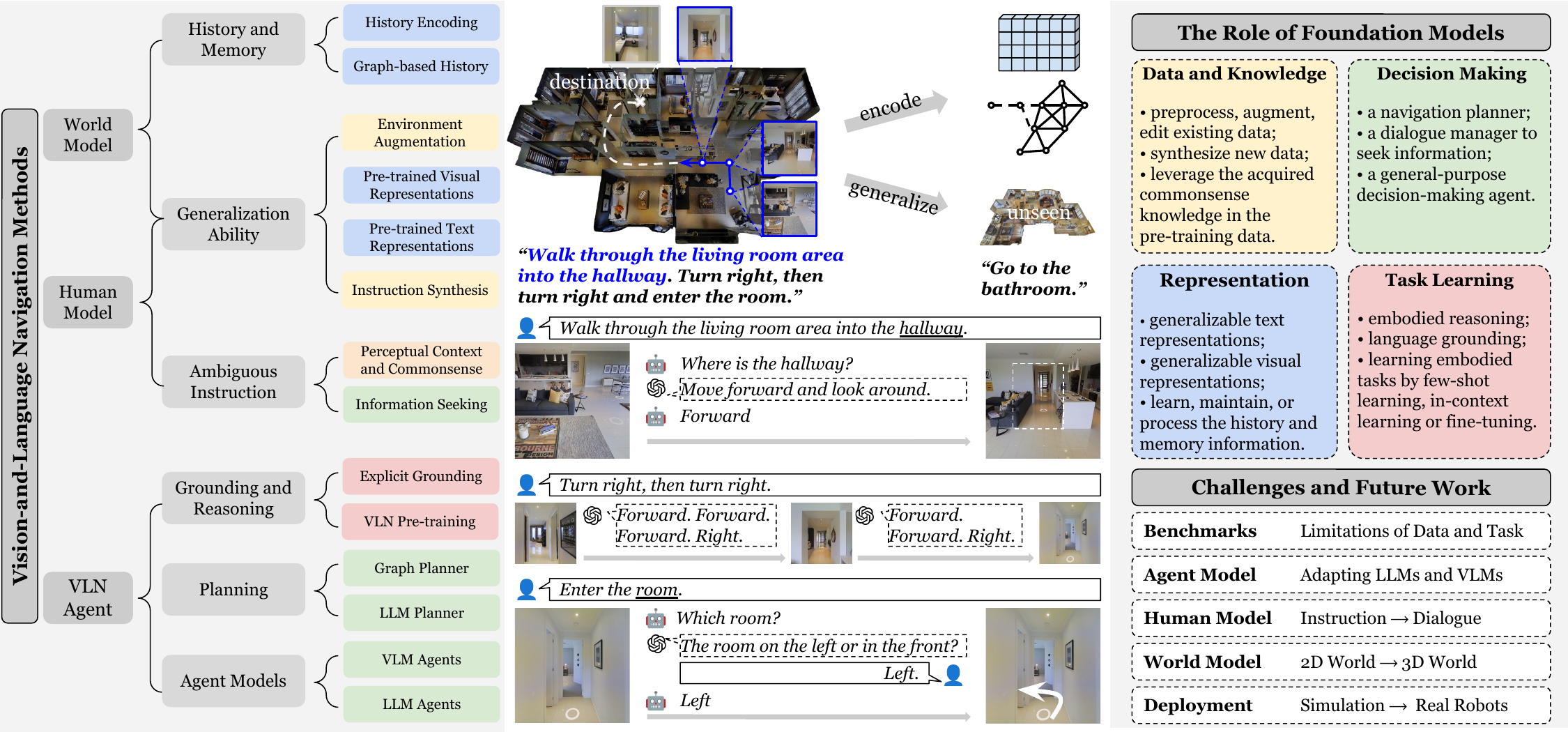}
    \caption{VLN challenges and solutions within the framework of world model, human model, and VLN agent. We discuss history and memory in the world model, ambiguous instructions in the human model, generalization ability in them both. For the VLN agent, we discuss methods for grounding and reasoning, planning, and adapting foundation models as agents. Depending on the role served by the foundation models, we categorize these methods into four types. Additionally, we discuss the potential future of the foundation model for the VLN task. 
   }
    \label{fig:classification}
\end{figure*}

Specifically, at each navigation step, the AI agents perceive the visual environment, receive language instructions from humans, and reason upon their representation of the world and humans to plan actions and efficiently complete navigation tasks. 
As shown in Figure~\ref{fig:world and agent}, a \textbf{\textit{world model}} is an abstraction that agents maintain to understand the external environment around them and how their actions change the world state~\citep{ha2018worldmodels,koh2021pathdreamer}.
This model is part of a broader \textbf{\textit{agent model}}, which also incorporates a \textbf{\textit{human model}} that interprets the instructions of its human partner, thereby informing the agent’s goals~\citep{andreas2022language,ma2023towards}.
To review the growing body of work in VLN and to understand the milestones achieved, we adopt a top-down approach to survey the field, focusing on fundamental challenges from three perspectives:
\begin{itemize}
    \item Learning a world model to represent the visual environment and generalize to unseen ones.
    \item Learning a human model to effectively interpret human intentions from grounded instructions. 
    \item Learning a VLN agent that leverages its world and human model to ground language, communicate, reason, and plan, enabling it to navigate environments as instructed. 
\end{itemize}
We present a hierarchical and fine-grained taxonomy in Figure~\ref{fig:classification} to discuss challenges, solutions, and future directions based on foundation models for each model.
To organize this survey, we start with a brief overview of the background and related research efforts as well as the available benchmarks in this field (\S\ref{sec:background}). 
We structure the review around how the proposed methods have addressed the three key challenges described above: world model (\S\ref{sec:world}), human model (\S\ref{sec:human}), and VLN agent (\S\ref{sec:agent}).
Finally, we discuss the current challenges and future research opportunities, particularly in light of the rise of foundation models (\S\ref{sec:future}).

\section{Background and Task Formulations}
\label{sec:background}

In this section, we discuss the background, clarify the scope of this survey, define the VLN problem, and briefly overview the benchmarks.

\begin{table*}[t]
\centering
    \scalebox{0.8}{
    \begingroup
    \setlength{\tabcolsep}{3.2pt}
    \renewcommand{\arraystretch}{0.8}
    \hspace*{-10pt}
    \begin{tabular}{ccccccccccc}
    \toprule
    \multirow{2}{*}{\textbf{Name}} & \multicolumn{2}{c}{\textbf{World}}      & \multicolumn{3}{c}{\textbf{Human}}               & \multicolumn{3}{c}{\textbf{VLN Agent}}                                      & \multicolumn{2}{c}{\textbf{Dataset}}        \\
    \cmidrule(r){2-3}\cmidrule(r){4-6}\cmidrule(r){7-9}\cmidrule(r){10-11}
                                   & \textbf{Domain} & \textbf{Environment}  & \textbf{Turn} & \textbf{Format} & \textbf{Gran.} & \textbf{Type} & \textbf{Act. Sp.} & {\textbf{Other}} & \textbf{Text}        & \textbf{Route}       \\
    \cmidrule(r){1-1}\cmidrule(r){2-3}\cmidrule(r){4-6}\cmidrule(r){7-9}\cmidrule(r){10-11}
    LANI/CHAI~\shortcite{misra2018mapping}                      & Indoors         & CHALET                & Single        & Multi Instr     & A              & -             & Disc              & {Mani}                & H                    & H                    \\
    R2R~\shortcite{mattersim}                            & Indoors         & Matterport3D          & Single        & Multi Instr     & A              & Robot         & Graph             &                                         & H                    & P                    \\
    R4R~\shortcite{jain2019stay}                            & Indoors         & Matterport3D          & Single        & Multi Instr     & A              & Robot         & Graph             &                                         & H                    & P                    \\
    RxR~\shortcite{ku2020room}                            & Indoors         & Matterport3D          & Single        & Multi Instr     & A              & Robot         & Graph             &                                         & H                    & P                    \\
    SOON~\shortcite{zhu2021soon}                           & Indoors         & Matterport3D          & Single        & Multi Instr     & G              & Robot         & Graph             &                                         & H                    & P                    \\
    REVERIE~\shortcite{qi2020reverie}                        & Indoors         & Matterport3D          & Single        & Multi Instr     & A, G           & Robot         & Graph             & {Detect}              & H                    & P                    \\
    VNLA~\shortcite{nguyen2019vision}                           & Indoors         & Matterport3D          & Multi         & Multi Instr     & A, G           & Robot         & Graph             &                                         & T                    & P                    \\
    HANNA~\shortcite{nguyen2019help}                          & Indoors         & Matterport3D          & Multi         & Multi Instr     & A, G           & Robot         & Graph             &                                         & H                    & P                    \\
    CVDN~\shortcite{thomason2020vision}                           & Indoors         & Matterport3D          & Multi         & Restricted      & A              & Robot         & Graph             &                                         & H                    & H                    \\
    VLN-CE~\shortcite{vlnce}                         & Indoors         & Habitat, Matterport3D & Single        & Multi Instr     & A              & Robot         & Disc              &                                         & H                    & P                    \\
    Robo-VLN~\shortcite{irshad2021hierarchical}                       & Indoors         & Habitat, Matterport3D & Single        & Multi Instr     & A              & Robot         & Cont              &                                         & H                    & P                    \\
    RobotSlang~\shortcite{banerjee2021robotslang}                     & Indoors         & Real                  & Multi         & Freeform        & A              & Robot         & Disc              &                                         & H                    & P                    \\
    % III~\shortcite{}                            & Indoors         & Playroom              & Single        & Single Instr    & A              & -             & Disc                & {Mani}                & H                    & P                    \\
    ALFRED~\shortcite{shridhar2020alfred}                         & Indoors         & AI2-THOR              & Single        & Multi Instr.    & A, G           & Robot         &Disc                & {Mani}                & H                    & P                    \\
    TEACh~\shortcite{padmakumar2022teach}                          & Indoors         & AI2-THOR              & Multi         & Freeform        & A, G           & Robot         &Disc                & {Mani}                & H                    & H                    \\
    DialFRED~\shortcite{gao2022dialfred}                       & Indoors         & AI2-THOR              & Multi         & Restricted      & A, G           & Robot         & Disc                & {Mani}                & H, T                 & P                    \\
    TouchDown~\shortcite{chen2019touchdown}                      & Outdoors        & Google Street View                   & Single        & Multi Instr     & A              & -             & Graph             &                                         & H                    & P                    \\
    Street Nav~\shortcite{hermann2020learning}                     & Outdoors        & Google Street View                   & Multi         & Multi Instr     & A              & -             & Disc                &                                         & T                    & P                    \\
    Talk2Nav~\shortcite{vasudevan2021talk2nav}                       & Outdoors        & Google Street View                   & Single        & Multi Instr     & A, G           & -             & Disc                &                                         & H                    & P                    \\
    TtW~\shortcite{de2018talk}                            & Outdoors        & Real                  & Multi         & Freeform        & A, G           & -             & Disc              &                                         & H                    & H                    \\
    LCSD~\shortcite{sriram2019talk}                           & Outdoors        & CARLA                 & Single        & Multi Instr     & A              & Driving       & Disc                &                                         & H                    & P                    \\
    CDNLI~\shortcite{roh2020conditional}                          & Outdoors        & CARLA                 & Multi         & Multi Instr     & A, G           & Driving       & Cont              &                                         & H, T                 & H                    \\
    SDN~\shortcite{ma2022dorothie}                            & Outdoors        & CARLA                 & Multi         & Freeform        & A, G           & Driving       & Disc, Cont        &                                         & H                    & H                    \\
    AerialVLN~\shortcite{liu2023aerialvln}                      & Outdoors        & AirSim                & Single        & Multi Instr     & A, G           & Aerial        & Disc              &                                         & H                    & H                    \\
    ANDH~\shortcite{fan2023aerial}                           & Outdoors        & xView                 & Multi         & Freeform        & A, G           & Aerial        & Disc              &                                         & H                    & H                   \\
    \bottomrule
    \end{tabular}
    \endgroup}
    % \vspace*{-5pt}
    \caption{A summary of existing VLN benchmarks, taxonomized based on several key aspects: the \textbf{world} in which navigation occurs, the type of \textbf{human} interaction involved, the action space and tasks assigned to the \textbf{VLN agent}, and the methods of \textbf{dataset} collection. For the \textbf{world}, we consider their \textbf{domain} (either indoors or outdoors) and the \textbf{environment}. For the \textbf{human}, we consider their \textbf{turns of interaction} (either single or multiple turn), the \textbf{format of communication} (freeform dialogue, restricted dialogue, or multiple instructions), and the \textbf{language granularity} (action-directed and goal-directed). For the \textbf{VLN agent}, we consider their \textbf{agent types} (e.g., household robot, autonomous driving vehicles, or autonomous aerial vehicles), their \textbf{action space} (graph-based, discrete or continuous), and \textbf{other additional tasks} (manipulation and object detection). For \textbf{dataset collection}, we consider the \textbf{text collection} (by human or templated) and the \textbf{route demonstrations} (by human or planner).}
    \label{tab:dataset}
\end{table*}

\subsection{Cognitive Underpinnings of VLN}

Humans and other navigational animals demonstrate early understanding and strategies for navigating their environments~\citep{rodrigo2002navigational,brand2015crawling,lingwood2018using}.
For example, \citet{gallistel1990organization} describes two basic mechanisms: \textit{piloting}, which involves environmental landmarks and computes distances and angles; and \textit{path integration}, which calculates displacement and orientation changes through self-motion sensing.
Central to understanding spatial navigation is the \textit{cognitive map hypothesis}, suggesting that the brain forms a unified spatial representation to support memory and guide navigation~\citep{epstein2017cognitive,bellmund2018navigating}.
For instance, \citet{tolman1948cognitive} observed that rats could adopt the correct novel path when familiar paths are blocked and landmarks are absent.
Neuroscientists also discovered hippocampal place cells, indicating a spatial coordinate system that encodes landmarks and goals allocentrically~\citep{o1971hippocampus,o1978hippocampus}.
Recent studies also propose non-Euclidean representations, \textit{e.g.}, \textit{cognitive graphs}, which illustrate the complexity of how we represent spatial knowledge of the world~\citep{warren2019non,ericson2020probing}. 
While visual and auditory perceptions are obviously integral to spatial representation~\citep{klatzky2006cognitive}, our linguistic skills and spatial cognition are also closely intertwined~\citep{pruden2011children}.
For instance, researchers have shown that understanding different aspects of spatial language can help with space-related tasks~\citep{pyers2010evidence}, and that language influences how children interact with space by assisting them to recognize the importance of landmarks in identifying locations~\citep{shusterman2011cognitive}.
Studying VLN not only enhances the development of embodied AI that follows human instructions in visual environments, but also deepens our understanding of how cognitive agents develop navigation skills, adapt to different environments, and how language use is connected to visual perceptions and actions.

\subsection{Relevant Tasks and Scope of the Survey}

Following natural language navigation instructions has traditionally been modeled using symbolic world representations such as maps~\citep{anderson1991hcrc,macmahon2006walk,paz2019run}. 
However, our survey focuses on models that employ visual environments and address the challenges of multimodal understanding and grounding.
Likewise, we redirect readers to extensive surveys on visual navigation~\citep{zhu2021deep,zhang2022survey,zhu2022integrity} and mobile robot navigation~\citep{gul2019comprehensive,crespo2020semantic,moller2021survey}, which concentrate on visual perception and physical embodiment. However, these studies provide minimal discussions on the role of language in navigation tasks.
While we inevitably extend our discussions of VLN to encompass areas beyond navigation, such as mobile manipulation and dialogue, our primary focus remains on navigational tasks, for which we provide a detailed literature review.
Besides, unlike previous VLN surveys~\citep{gu2022vision,park2023visual,wu2024vision}, which offer a bottom-up summary focusing on benchmarks and modeling innovations, our survey adopts a top-down approach, and uses the roles of foundation models to categorize the research efforts into three fundamental challenges from the aspects of the world model, the human model, and the VLN agent. 
Note that this survey concentrates on frontier methods associated with the rise of foundation models. 
Thus, we point to the earlier generation of models (\textit{e.g.}, LSTM-based methods) very briefly at the beginning of each section to motivate our discussions.

\subsection{VLN Task Formulations and Benchmarks}
\label{sec:dataset}

\paragraph{VLN Task Definition.} 

A typical VLN agent receives a (sequence of) language instruction(s) from human instructors at a designated position.
The agent navigates through the environment using an egocentric visual perspective. 
By following the instructions, its task is to generate a trajectory over a sequence of discrete views or lower-level actions and control (\textit{e.g.}, FORWARD $0.25$ meter) to reach the destination, which is considered successful if the agent arrives within a specified distance (\textit{e.g.}, $3$ meters) from the destination.
Besides, the agent may exchange information with the instructor during navigation, either by requesting help or engaging in freeform language communication. 
Additionally, there has been an increasing expectation for VLN agents to integrate additional tasks such as manipulation~\citep{shridhar2020alfred} and object detection~\citep{qi2020reverie}, along with navigation.

\paragraph{Benchmarks.} Unlike other multimodal tasks such as VQA, which have a relatively fixed task definition and format, VLN encompasses a wide range of benchmarks and task formulations. 
These distinctions introduce unique challenges in addressing the broader VLN task and must be clearly understood as prerequisites for developing effective methods with appropriate foundation models.
As is summarized in Table~\ref{tab:dataset}, existing VLN benchmarks can be taxonomized based on several key aspects in the LAW framework: 
(1) the \textbf{world} where navigation occurs, including the domain (indoors or outdoors) and the specifics of the environment.
(2) the type of \textbf{human} interaction involved, including the interaction turns (single or multiple), communication format (freeform dialogue, restricted dialogue, or multiple instructions), and language granularity (action-directed or goal-directed).
(3) the \textbf{VLN agent}, including its types (\textit{e.g.}, household robots, autonomous driving vehicles, or autonomous aerial vehicles), action space (graph-based, discrete, or continuous), and additional tasks (manipulation and object detection).
(4) the \textbf{dataset} collection, including text collection method (human-generated or templated) and route demonstrations (human-performed or planner-generated).
Representatively, \citet{mattersim} create the Room-to-Room (R2R) dataset based on the Matterport3D simulator~\citep{chang2017matterport3d}, where an agent needs to follow fine-grained navigation instructions to reach the goal. 
Room-across-Room (RxR)~\citep{ku2020room} is a multilingual variation, including English, Hindi, and Telugu instructions. 
It offers a larger sample size and provides time-aligned instructions for virtual poses, enriching the task's linguistic and spatial information. 
Matterport3D allows VLN agents to operate in a \textit{discrete environment} and rely on pre-defined connectivity graphs for navigation, where agents travel on the graph by teleportation between adjacent nodes, referred to as VLN-DE.
To make the simplified setting more realistic, \citet{vlnce,li2022reve,irshad2021hierarchical} propose VLN in \textit{continuous environments} (VLN-CE) by transferring discrete R2R paths to continuous spaces~\citep{habitat}.
Robo-VLN~\citep{irshad2021hierarchical} further narrows the sim-to-real gap by introducing VLN with continuous action spaces that are more realistic in robotics settings.
Recent VLN benchmarks have undergone several design changes and expectations, which we discuss in \S~\ref{sec:future}.

\paragraph{Evaluation Metrics.} 
Three main metrics have been employed to evaluate navigation wayfinding performance~\citep{mattersim}: 
(1) \textit{Navigation Error (NE)}, the mean of the shortest path distance between the agent's final position and the goal destination; 
(2) \textit{Success Rate (SR)}, the percentage of the final position being close enough to the goal destination; and
(3) \textit{Success Rate Weighted Path Length (SPL)}, which normalizes success rate by trajectory length to balance both the success rate in reaching the correct destination and the efficiency of the path. 
Some other metrics are used to measure the faithfulness of instruction following and the fidelity between the predicted and the ground-truth trajectory, for example:
(4) \textit{Coverage Weighted by Length Score (CLS)}~\citep{jain2019stay}
 measures how closely an agent's trajectory follows the reference path. It balances two key aspects of the agent's performance: the extent of coverage of the reference path and the efficiency of the agent's navigation by considering the length score;
(5) \textit{Normalized Dynamic Time Warping (nDTW)}~\citep{ndtw}, which penalizes deviations from the ground-truth trajectories; and
(6) \textit{Normalized Dynamic Time Warping Weighted by Success Rate (sDTW)}~\citep{ndtw}, which penalizes deviations from the ground-truth trajectories and also considers the success rate.

\subsection{Foundation Models}
Foundation models are trained on large-scale datasets, which show strong generalization capability for a wide range of downstream applications.  
Text-only foundation models, such as pre-trained language models like BERT~\citep{devlin2018bert} and GPT-3~\citep{brown2020language}, have revolutionized the field of NLP by setting new benchmarks for tasks like text generation, translation, and understanding. 
Building on the success of these models, vision-language (VL) foundation models, like LXMERT~\citep{tan2019lxmert}, CLIP~\citep{clip} and GPT-4~\citep{openai2023gpt4}, have expanded the paradigm to multimodal learning by integrating both visual and textual data, proving particularly impactful in various VL applications~\citep{li2019visualbert, ramesh2021zero, alayrac2022flamingo,rvlnbert, zhang2025common, cheng2024shine, kamali2023syntax}. 
For a more comprehensive overview of foundation models and their applications, we encourage readers to refer to existing survey papers such as~\citet{bommasani2021opportunities}, \citet{du2022survey}, and \citet{zhou2023comprehensive}.

\vspace*{-3pt}
\section{World Model: Learning and Representing the Visual Environments}
\label{sec:world}

\vspace*{-2pt}
A world model helps the VLN agent to understand their surrounding environments, predict how their actions would change the world state, and align their perception and actions with language instructions. 
Two challenges have been highlighted in existing work about learning a world model: encoding the visual history of observations within the current episode as memory, and achieving generalization to unseen environments.

\subsection{History and Memory}

Different from other vision-language tasks like Visual Question Answering (VQA)~\citep{VQA}, Visual Entailment~\citep{xie2019visual}, the VLN agent needs to incorporate the history information of past actions and observations into its current step’s input to determine the action rather than solely considering the image and text in a single step.
Prior to employing the foundation models in VLN, LSTM hidden states served as an implicit memory supporting agents' decision-making during navigation, and researchers further design different attention mechanisms~\citep{tan2019learning, wang2019reinforced} or auxiliary tasks~\citep{ma2019self, zhu2020vision} to improve the alignment between the encoded history and instructions.

\paragraph{History Encoding.}

Different techniques have been proposed to encode navigation history using foundation models. A multi-modal Transformer is built upon encoded instructions and navigation history for decision-making, which is usually initialized from a model pre-trained on in-domain instruction-trajectory data like Prevalent~\citep{prevalent}. Some approaches encode the navigation history in recurrently updated state tokens. \citet{rvlnbert} proposes to utilize a single \texttt{[CLS]} token from last step for encoding the history information, while \citet{mtvm} introduces a variable-length memory framework that stores multiple action activations from previous steps in a memory bank as the history encoding. Despite their effectiveness, these methods are limited by the need for step-by-step token updates, making it challenging to efficiently retrieve history encodings at arbitrary steps in the navigation trajectory, which can hinder scalability in pre-training. Another line of work directly encodes navigation history as a sequence with multi-modal Transformer. Among them, \citet{ET} encodes single-view images for each step in a trajectory. \citet{hamt} further proposes a panorama encoder to encode the panoramic visual observation at each time step, followed by a history encoder to encode all the past observations. This hierarchical design separately processes the spatial relationship in a panoramic view and the temporal dynamics across panoramas in the navigation history. Besides, this method eliminates the dependency on recurrently updated state tokens for history encoding, facilitating efficient and large-scale pre-training on instruction-path pairs. Follow-up research replaces the panorama encoder with mean pooling of images~\citep{kamath2023new} or front-view image encoding~\citep{hop}, both maintaining effective navigation performance. With the advent of LLM-based navigation agents, some works~\citep{zhou2024navgpt} focus on converting the visual environment into textual descriptions, and explaining the world with text became the trend. The navigation history is then encoded as a sequence of these image descriptions, along with relative spatial information such as heading, elevation, and distance. HELPER~\citep{sarch2023open} designs an external memory of language-program pairs that parses the free-form human-robot dialogue into action programs through retrieval-augmented LLM prompting.

\paragraph{Graph-based History.} 
Another line of research enhances the navigation history modeling with graph information. For example, some of these techniques utilize a structured Transformer encoder to capture the geometric cues in the environment~\citep{chen2022think, deng2020evolving, wang2023dual, zhou2023tree, su2023target, zheng2023esceme, wang2021structured, chen2021topological, zhu2021soon}. In addition to the topological graph used in encoding, many works propose to include the top-down view information (\textit{e.g.}, grid map~\citep{wang2023gridmm, liu2023bird}, semantic map~\citep{hong2023learning, huang2023visual, cm2, anderson2019chasing, WS-MGMap, irshad2022semantically}, local metrics map~\citep{an2023bevbert}), and local neighborhood map~\citep{gopinathan2023near} in modeling the observation history during navigation. Recent advances in LLM-based navigation agents have introduced innovative approaches to memory construction using maps. For instance, \citet{chen2024mapgpt} proposes a novel map-guided GPT-based agent that utilizes a linguistical-formed map to store and manage topological graph information. MC-GPT~\citep{zhan2024mc} introduces a topological map as the memory structure to record information about viewpoints, objects, and their spatial relationships. 

\subsection{Generalization across Environments}
One main challenge in the VLN is learning from limited available environments and generalizing to new and unseen environments. Many works demonstrate that learning from semantic segmentation features~\citep{zhang2020diagnosing}, dropout information in the environment during training~\citep{tan2019learning}, and maximizing the similarity between semantically-aligned image pairs from different environments~\citep{li2022clear} improve agents' generalization performance to unseen environments. These observations suggest the need to learn from large-scale environment data to avoid overfitting to training environments.
Next, we discuss how existing works collect new environment data, and utilize it in training.

\paragraph{Pre-trained Visual Representations.} 
Most works obtain vision representations from ResNet pre-trained on ImageNet~\citep{mattersim, tan2019learning}. 
\citet{shen2021much} replace ResNet with the CLIP visual encoder~\citep{clip}, which is pre-trained with contrastive loss between image-text pairs and naturally better aligns the image with the instructions, boosting the VLN performance. 
\citet{wang2022internvideo} further explores transferring vision representation learned from video data for VLN task, suggesting that temporal information learned from video is crucial for navigation.

\paragraph{Environment Augmentation.} One main line of research focuses on augmenting the navigation environment with auto-generated synthetic data. EnvEdit~\citep{li2022envedit}, EnvMix~\citep{liu2021vision}, KED~\citep{zhu2023vision}, and FDA~\citep{he2024frequency} generate synthetic data by changing the existing environments from Matterport3D. Specifically, they mix up rooms from different environments, change the appearance and style of the environments, and interpolate high-frequency features with the environments. 
Pathdreamer~\citep{koh2021pathdreamer} and SE3DS~\citep{koh2023simple} further synthesize the environments in future steps given current observations and explore utilizing the synthesis view as augmented data for VLN training.

The learning paradigm from the collected environments has changed with the advances in foundation models. Prior to the prevalence of pre-training in foundation models, most works directly augment the training environment with the auto-collected new environments and fine-tune a LSTM-based VLN agent~\citep{li2022envedit, liu2021vision, koh2021pathdreamer, koh2023simple, zhu2023vision}.
As pre-training has been demonstrated to be crucial for foundation models, it has also become a standard practice in VLN to learn from collected environments during the pre-training stage~\citep{li2024panogen, kamath2023new, chen2022learning, wang2023scaling, lin2023learning, airbert, he2024frequency}. Large-scale pre-training with augmented in-domain data has become crucial in bridging the gap between agents' and humans' performance. The in-domain pre-trained multi-modal transformer has been proven to be more effective than the multi-modal Transformer initialized from VLMs, like Oscar~\citep{li2020oscar} and LXMERT.

\section{Human Model: Interpreting and Communicating with Humans}
\label{sec:human}
In addition to learning and modeling the world, VLN agents  need a human model that comprehends human-provided natural language instructions per situation to complete navigation tasks. 
There are two main challenges: resolving ambiguity and generalization of grounded instructions in different visual environments.

\subsection{Ambiguous Instructions}
Ambiguous instructions mainly arise in single-turn navigation scenarios, where the agent follows an initial instruction without further human interaction for clarification. 
These instructions lack the flexibility to train the agent to adapt its language understanding and visual perception to the dynamic environments. For instance, instructions may contain landmarks invisible at the current view or indistinguishable landmarks visible from multiple views~\citep{zhang2023vln}.
The issue of ambiguous instructions is barely addressed before the application of foundational models to VLN. Although LEO~\citep{xia2020multi} attempts to aggregate multiple instructions to describe the same trajectory from different perspectives, it still relies on human-annotated instructions. 
However, comprehensive perceptual context and commonsense knowledge from foundational models enable the agent to interpret ambiguous instructions using external knowledge, as well as seek assistance from other human models. 

\paragraph{Perceptual Context and Commonsense Knowledge.} 
Large-scale cross-modal pre-trained models like CLIP are capable of matching visual semantics with text. 
This enables the VLN agent to utilize information from the visual objects and their states in the current perception to resolve ambiguity, especially in single-turn navigation scenarios.
For example, VLN-Trans~\citep{zhang2023vln} constructs easy-to-follow sub-instructions with visible and distinctive objects obtained from CLIP to pre-train a Translator that converts original ambiguous instructions into easily understandable sub-instruction representations. 
LANA+~\citep{wang2023learning} leverages CLIP to query a text list of landmark semantic tags with the visual panoramic observations, and
selects the top-ranked retrieved textual cues as representations of the salient landmarks to follow. KERM~\citep{li2023kerm} proposes a knowledge-enhanced reasoning model to retrieve facts where knowledge is described by language descriptions for the navigation views.
NavHint~\citep{zhang2024navhint} constructs a hint dataset, providing detailed visual descriptions to help the VLN agent build a comprehensive understanding of the visual environment rather than focusing solely on the objects mentioned in the instructions.
On the other hand, the commonsense reasoning ability of LLMs can be used to clarify or correct ambiguous landmarks in the instructions, and break instructions into actionable items.
For example, \citet{lin2024correctable} use LLMs to provide commonsense about open-world landmark co-occurrences and conduct CLIP-driven landmark discovery accordingly.
SayCan~\citep{ahn2022can} breaks an instruction into a ranked list of pre-defined admissible actions and combines them with an affordance function that assigns higher weights to the objects appearing in the current scene.

\paragraph{Information Seeking.}
While ambiguous instructions can be resolved based on visual perception and situational context, another more direct approach is to seek help from the communication partner, i.e., the human speakers who generate the instructions~\citep{nguyen2019help,paul2022avlen}.
There are three key challenges in this line of work: (1) deciding when to ask for help~\citep{chi2020just}; (2) generating information-seeking questions, \textit{e.g.}, next action, objects, and directions~\citep{roman2020rmm,singh2022ask4help}; (3) developing an oracle that provides the queried information, which could be either real humans~\citep{singh2022ask4help}, rules and templates~\citep{gao2022dialfred}, or neural models~\citep{nguyen2019help}.
LLMs and VLMs could potentially fit two roles in this framework, either as information-seeking models, or as proxies for human helpers or information-providing models. 
Preliminary research has explored the use of LLMs as the information-seeking model, addressing determining both when and what to ask.
This is achieved with the help of techniques including conformal prediction (CP)~\citep{ren2023robots} or in-context learning (ICL)~\citep{chen2023asking}.
For the latter, foundation models play the role of a helper who has access to oracle information, such as the location of the destination and a map of the environment, which is not available to the task performer. Very recently, VLN-Copilot~\citep{qiao2024llm} enables agents to actively seek assistance when encountering confusion, with the LLM serving as a copilot to facilitate navigation.
\citet{fan2023r2h} demonstrate that GPT-3 can decompose ground-truth responses in the training data step-by-step, which helps in training an oracle model using a pre-trained SwinBert~\citep{lin2022swinbert} video-language model.
They also demonstrate large vision-language models like mPLUG-Owl~\citep{ye2023mplug} can serve as strong zero-shot oracles off the shelf. 
In addition, self-motivated communication agents have been developed~\citep{zhu2021self} by learning the confidence of the oracle to produce a positive answer, which enables a self-Q\&A manner where the oracle can be removed at inference time.

\subsection{Generalization of Grounded Instructions}
The limited scale and diversity of navigation data is another significant issue affecting the VLN agent's ability to comprehend various linguistic expressions and follow instructions effectively, particularly in unseen navigation environments. Although the language style itself has good generalization capability across seen and unseen environments~\citep{zhang2020diagnosing}, how to ground the instructions with the unseen environments is potentially a hard task given the limited scale of training instructions. Foundation models help address these issues through both pre-trained representations and instruction generation for data augmentation.

\paragraph{Pre-trained Text Representations.} Before the foundation models, many works rely on text encoders, such as LSTM, to represent text instructions~\citep{mattersim, tan2019learning}. 
The foundation models significantly enhance the VLN agent's language generalization ability through pre-trained representations.
For example, PRESS~\citep{press} fine-tunes the pre-trained language model BERT~\citep{devlin2018bert} to obtain text representations that generalize better to previously unseen instructions. 
The multi-modal Transformers~\citep{tan2019lxmert, lu2019vilbert} boost methods, such as VLN-BERT~\citep{bertvln} and PREVALENT~\citep{prevalent},  to obtain more generic vision-linguistic representations by pre-training on large-scale text-image pairs collected from the web. Airbert~\citep{guhur2021airbert} trains ViLBERT-like architecture to learn text representations from image-caption pairs collected from the Internet. CLEAR~\citep{li2022clear} learns cross-lingual language representations that capture the visual concepts behind the instruction. ProbES~\citep{liang2022visual} self-explores environments by sampling trajectories and automatically constructs the corresponding instruction by filling the instruction templates with movements and object phrases detected by CLIP. Additionally, it leverages prompt-based learning to facilitate fast adaptation of language embeddings.
NavGPT-2~\citep{zhou2024navgpttwo} explores leveraging vision-and-language representations from pre-trained VLMs (InstructBLIP~\citep{dai2024instructblip} with Flan-T5~\citep{chung2024scaling} or Vicuna~\citep{zheng2023judging}) to enhance policy learning for navigation and navigational reasoning.

\paragraph{Instruction Synthesis.} Another method to improve the agent's generalization ability is to synthesize more instructions. Early works employ the Speaker-Follower framework~\citep{fried2018speaker, tan2019learning, kurita2020generative, airbert} to train an offline speaker (instruction generator) using human-annotated instruction-trajectory pairs. It then generates new instructions based on sequences of panoramas along a given trajectory.
However,~\citet{zhao2021evaluation} observe that these generated instructions are low-quality and show a poor performance in human wayfinding evaluation.
Marky~\citep{wang2022less, kamath2023new} addresses this limitation using a multi-modal extension of the multilingual T5 model~\citep{xue2020mt5} with text-aligned visual landmark correspondences, achieving near-human quality on R2R-style paths in unseen environments. PASTS~\citep{Wang2023PASTSPS} introduces a progress-aware spatial-temporal Transformer speaker to better leverage the sequenced multiple vision and action features. SAS~\citep{gopinathan2024spatially} generates instructions with rich spatial information using semantic and structural cues from the environment. SRDF~\citep{wang2024bootstrapping} builds a strong instruction generator with iterative self-training.
Additionally, instead of training an offline instruction generator, some recent research~\citep{liang2022visual,lin2023learning, zhang2023vln, wang2023lana, Magassouba2021CrossMapTA} generates instructions while navigating. For instance, LANA~\citep{wang2023lana} introduces a language-capable navigation agent that not only executes navigation instructions but also provides route descriptions. 

\section{VLN Agent: Learning an Embodied Agent for Reasoning and Planning}
\label{sec:agent}
While the world and human models empower visual and language understanding abilities, VLN agents need to develop embodied reasoning and planning capabilities to support their decision-making.
From this perspective, we discuss two challenges: grounding and reasoning, and planning. 
We also explore the method of directly applying foundation models as the VLN agent backbone.

\subsection{Grounding and Reasoning}
Different from other VL tasks, such as VQA and Image Captioning, which primarily focus on static alignment between images and corresponding textual descriptions, the VLN agent needs to reason about spatial and temporal dynamics in the instructions and the environment based on its actions.
Specifically, the agent should consider previous actions, identify the part of the sub-instruction to execute, and ground the text to the visual environment to execute the action accordingly. 
Previous methods primarily rely on explicit semantic modeling or auxiliary task design to obtain such abilities.  However, pre-training with specially designed tasks has become the dominant approach with the advent of foundation models.

\paragraph{Explicit Semantic Grounding.} The previous efforts enhance the agent's grounding ability through explicit semantic modeling in both vision and language modalities, including modeling motions and landmarks~\citep{hong2020sub,he2021landmark,hong2020language,zhang2021towards,qi2020object}, utilizing syntactic information in the instruction~\citep{li2021improving}, as well as spatial relations~\citep{zhang2022explicit,an2021neighbor}.
Very few works~\citep{lin2023actional, zhan2024enhancing, wang2023dual} explore explicit grounding in the VLN agent with the foundation models. \citet{lin2023actional} proposes actional atomic-concept learning and map visual observations to faciliate multi-modal alignments.

\paragraph{Pre-training VLN Foundation Models.} 
Except for explicit semantic modeling, the previous research also enhances the agent's grounding ability through auxiliary reasoning tasks~\citep{ma2019self, wu2021multi, zhu2020vision, LAW, dou2022foam, kim2021ndh}. 
Such methods are less explored in VLN agents with foundation models, as their pre-training already provides a general understanding of spatial and temporal semantics prior to navigation.
Various pre-training methods with specially designed tasks have been proposed to improve the agent's grounding ability.
\citet{lin2021scene} introduce pre-training tasks specifically designed for scene and object grounding.
LOViS~\citep{lovis} formulates two specialized pre-training tasks to enhance orientation and visual information separately.
HOP~\citep{hop,hop+} introduces a history-and-order aware pre-training paradigm that emphasizes historical information and trajectory orders.
\citet{li2023improving} suggests that enhancing the agent with the ability to predict future view semantics helps the agent in longer path navigation performance.
\citet{DouGP23} design a masked path modeling objective to reconstruct the original path given a randomly masked sub-path.
\citet{cui2023grounded} propose entity-aware pre-training by predicting grounded entities and aligning them to text.

\subsection{Planning}
Dynamic planning enables VLN agents to adapt to environmental changes and improve navigation strategies on the fly. 
Alongside the graph-based planners that utilize global graph information to enhance local action spaces, the rise of foundational models, particularly LLMs, has brought LLM-based planners into the VLN field. These planners use LLMs' vast commonsense knowledge and advanced reasoning to create dynamic plans that improve decision-making.

\paragraph{Graph-based Planner.} 
Recent advancements in VLN emphasize enhancing navigational agents' planning capabilities through global graph information.
Among them, \citet{wang2021structured, chen2022think, deng2020evolving, zheng2023esceme} enhance the local navigation action spaces with global action steps from graph frontiers of visited nodes for better global planning. \citet{gao2023adaptive} further enhances navigation decision-making with high-level planning for zone selection and low-level planning for node selection. Moreover, \citet{liu2023bird} enriches the graph-frontier-based global and local action spaces with grid-level actions for more accurate action prediction. 
In continuous environments, \citet{krantz2021waypoint,hong2022bridging,anderson2021sim} adopt a hierarchical planning approach utilizing high-level action spaces instead of low-level ones by selecting a local waypoint from a predicted local navigability graph. 
CM2~\citep{cm2} facilitates trajectory planning by grounding instructions within a local map. Expanding on this strategy, \citet{an2023etpnav,an2023bevbert,wang2023gridmm,chang2023goat,wang2022find} construct a global topological graph or grid maps to facilitate map-based global planning. Additionally, \citet{wang2023dreamwalker,wang2024lookahead} predict multiple future waypoints using either a video prediction model or a neural radiance representation model to plan the best action based on the long-term effects of predicted candidate waypoints.

\paragraph{LLM-based Planner.} 
In parallel, some studies leverage common-sense knowledge from LLMs to generate text-based plans~\citep{huang2022language, huang2022inner}.
LLM-Planner~\citep{song2022llm} creates detailed plans composed of sub-goals, dynamically adjusting these plans in real-time by integrating detected objects according to predefined program patterns. 
Similarly, Mic~\citep{mic} and $A^2$Nav~\citep{a2vnav} specialize in breaking down navigation tasks into detailed textual instructions, with Mic generating step-by-step plans from both static and dynamic perspectives, while $A^2$Nav uses GPT-3 to parse instructions into actionable sub-tasks. ThinkBot~\citep{lu2023thinkbot} employs thought chain reasoning to generate missing actions with interactive objects. VL-Map~\citep{huang2023visual} decomposes navigation instructions into sequential, goal-related functions in code format using code-written LLMs (following the Code-as-Policy~\citep{liang2023code} framework) and utilizes a dynamically built, queryable map to guide the execution of these goals.
Additionally, SayNav~\citep{rajvanshi2023saynav} builds a 3D scene graph of the explored environment as input to LLMs for generating feasible and contextually appropriate high-level plans for the navigator.

\subsection{Foundation Models as VLN Agents}
The architecture of VLN agents has undergone significant transformations with the advent of foundation models.
Initially conceptualized by \citet{mattersim}, VLN agents were formulated within a Seq2Seq framework, employing an LSTM and an attention mechanism to model the interaction between vision and language modalities. 
With the advent of foundation models, the agent backend has transitioned from LSTM to Transformer and, more recently, to these large-scale pre-trained systems.

\paragraph{VLMs as Agents.} 
The mainstream methodology leverages single-stream VLMs as the core structure of VLN agents~\citep{rvlnbert, orist, soat, TD-STP}.
These models process inputs from language, vision, and historical tokens simultaneously at each time step.
It performs self-attention over these cross-modal tokens to capture the textual-visual correspondence, which is then used to infer the action probability. In the zero-shot VLN, CLIP-NAV~\citep{dorbala2022clip} utilizes CLIP to obtain natural language referring expressions that describe the target object and make sequential navigational decisions.
VLN-CE agents~\citep{vlnce} differentiate themselves from the VLN-DE~\citep{mattersim} agents by their action space, executing low-level controls in the continuous environment instead of graph-based high-level actions of view selection. Despite early works~\citep{vlnce,LAW} utilizing LSTM to infer low-level actions, the introduction of waypoint predictors has allowed to transfer methods from DE to CE~\citep{krantz2021waypoint,krantz2022sim,hong2022bridging,anderson2021sim,an20221st, zhang2024narrowing}. 
All these methods use a waypoint predictor to obtain a local navigability graph, allowing foundation models in DE to adapt to the continuous environment. In particular, the waypoint detection process primarily involves using visual observations (\textit{e.g.}, panoramic RGBD images) to predict navigable candidate adjacent waypoints from the agent's current position as possible targets. Given the predicted waypoints, the agent selects one as the current destination.

\paragraph{LLMs as Agents.}  
Since LLMs have powerful reasoning ability and semantic abstraction of the world, and also show strong generalization ability in unknown large-scale environments, recent research in VLN has started to directly employ LLMs as agents to complete navigation. 
Typically, visual observations are converted into textual descriptions and fed into the LLM along with instructions, which then perform action predictions.
Innovations such as NavGPT~\citep{zhou2023navgpt} and MapGPT~\citep{chen2024mapgpt} demonstrate the feasibility of zero-shot navigation, with NavGPT autonomously generating actions using GPT-4 and MapGPT converting topological maps into global exploration hints. 
DiscussNav~\citep{long2023discuss} extends this approach by deploying multiple domain-specific VLN experts to automate and reduce human involvement in navigation tasks. It includes Instruction Analysis Experts, Vision Perception Experts, Completion Estimation Experts, and Decision Testing Experts. The use of multiple domain-specific VLN experts distributes tasks among specialized agents, reducing the burden on a single model and allowing for optimized, task-specific processing. This multi-expert approach enhances robustness, transparency, and overall performance by leveraging the collective strengths of multiple large models.
MC-GPT \citep{zhan2024mc} employs memory topology maps and human navigation examples to diversify strategies, while InstructNav \citep{long2024instructnavzeroshotgenericinstruction} breaks navigation into sub-tasks with multi-sourced value maps for effective execution. 
In contrast to zero-shot usage, some works~\citep{zheng2024towards, zhang2024navid, pan2024langnav} fine-tune LLMs to address the embodied navigation tasks effectively.
Some studies have incorporated the Chain-of-Thought (CoT)~\citep{cot} reasoning mechanism to improve the reasoning process. 
NavCoT~\citep{lin2024navcot} transforms LLMs into a world model and navigational reasoning agent, streamlining decisions by simulating future environments. 
This demonstrates the flexibility and practical potential of fine-tuned language models in both simulation and real-world scenarios, marking a significant advancement over traditional applications.
\section{Challenges and Future Directions}
\label{sec:future}

While foundation models have enabled novel solutions to VLN, several limitations remain under-explored, and new challenges arise.
In this section, we outline the challenges and future direction of VLN from the perspectives of benchmarks, the world model, the human model, the agent model, and real robot deployment.

% \vspace*{-5pt}
\paragraph{Benchmarks: Limitations of Data and Task.} 
The current VLN datasets have limitations regarding quality, diversity, bias, and scalability.
For example, in the R2R dataset, the instruction-trajectory pairs are biased to the shortest path, which may not accurately represent real-world navigation scenarios. 
We discuss the trends and recommendations on how VLN benchmarks can be improved.
\vspace*{-5pt}
\begin{itemize}[leftmargin=*]
\setlength\itemsep{-0.25em}

\item  \textit{Unified and Realistic Tasks and Platforms.} Establishing robust benchmarks and ensuring reproducibility are crucial for evaluating VLN in real-world settings. Real-world variability necessitates comprehensive benchmarks reflecting navigation challenges. A universal sim-to-real evaluation platform, like OVMM~\citep{yenamandra2024ovmm}, is needed for standardized testing across simulated and real-world settings. In addition, the tasks and activities should be realistic and designed originated from human needs. For instance, BEHAVIOR-1K~\citep{li2024behavior} presents a benchmark of everyday household activities in virtual, interactive, and ecological environment to address the demands for diversity and realism.

\item \textit{Dynamic Environment.} Real-world environments are inherently complex and dynamic, with moving objects, people, and variations like lighting and weather presenting unexpected situations~\citep{ma2022dorothie}. These factors disrupt the visual perception of navigation systems and make maintaining reliable performance difficult. Recent efforts like HAZARD~\citep{zhouhazard}, Habitat 3.0~\citep{puighabitat}, and HA-VLN~\citep{li2024havln} consider dynamic environments and provide a good starting point.

\item \textit{Indoors to Outdoors.}
VLN agents navigating in outdoor environments, \textit{e.g.}, autonomous driving and aerial vehicles, also start to get more attention~\citep{vasudevan2021talk2nav,li2024vln}, with various language-guided datasets~\citep{sriram2019talk,ma2022dorothie} developed.
Early studies have attempted to involve LLMs in these tasks, either with prompt engineering~\citep{shah2023lm,sha2023languagempc,wen2023dilu}, or by fine-tuning LLMs to predict the next action or plan future trajectories~\citep{chen2023driving,mao2023gpt}.
To adapt off-the-shelf VLMs to these outdoor navigation domains, real-world driving videos~\citep{xu2023drivegpt4,yuan2024rag}, simulated driving data~\citep{wang2023drivemlm,shao2023lmdrive} and them both~\citep{sima2023drivelm,huang2024drivlme} have been utilized for instruction tuning so that these foundation models learn to predict future throttle and steering angles. 
Additional reasoning and planning modules have also been integrated into foundation model driving agents~\citep{huang2024drivlme,tian2024drivevlm}.
We refer the readers to surveys and position papers for a detailed review~\citep{li2023towards,cui2024survey,gao2024survey,yan2024forging}.
\end{itemize}

\paragraph{World Model: From 2D to 3D.}
Building effective world representations is a central research theme in embodied perception, reasoning, and planning.
VLN is fundamentally a 3D task, where the agent perceives the real-world environment in 3D.
Although the current research represents the world with strong and generic 2D representations, they fall short of spatial language understanding in the 3D world~\citep{zhang2024vision}.
Many explicit 3D representations are developed in prior work, including various semantic SLAMs and volumetric representation~\citep{chaplot2020object,min2021film,saha2022modular,blukis2022persistent,zhang2022danli,liu2024volumetric}, depth information~\citep{an2023bevbert}, Bird's-Eye-View representations like grid map~\citep{wang2023gridmm, liu2023bird}, and local metrics map~\citep{an2023bevbert}. These representations are limited because they reduce the object set to a closed set, making them inadequate for open-vocabulary settings with natural language.
Several studies develop queryable map/scene representations by integrating multi-view image features captured from CLIP into 3D voxel grids~\citep{jatavallabhula2023conceptfusion,chang2023context} or top-down feature maps~\citep{huang2023visual,chen2022nlmapsaycan}, as well as utilizing scene graphs~\citep{rana2023sayplan,gu2023conceptgraphs} to represent spatial relationships. 
However, adapting 3D representations learned from large-scale data for VLN agents to better perceive the 3D environment is still under exploration. 
The recent rise of 3D foundation models~\citep{3dllm, huang2023embodied, chen2024bridging, chen2024sam}, including 3D reconstruction models~\citep{hong2024lrm} and 3D multimodal representations~\citep{yang2024llmgrounder, zhang2024spartun3d, zhang2024tamm}, can be crucial for VLN. 

\paragraph{Human Model: From Instruction to Dialogue.}
Previous efforts predominantly adopt either a speaker-listener paradigm or restricted QA dialogue~\citep{thomason2020vision,gao2022dialfred} that only allows the agent to ask for help.
Recently, there has been a surge in new benchmarks featuring open-ended dialogue instructions~\citep{de2018talk,banerjee2021robotslang,padmakumar2022teach,ma2022dorothie,fan2023aerial}, supporting fully free-form communication where agents can ask, propose, explain, suggest, clarify, and negotiate even in ambiguous or confusing scenarios.
Still, current approaches rely on rule-based dialogue templates to tackle these complexities~\citep{zhang2023seagull,parekh2023emma,gu2023slugjarvis}, though they might feature a foundation model component.
\citet{huang2024drivlme} perform conversational tuning on a video-language model using human-human dialogue data paired with simulated navigation videos, showcasing enhanced dialogue generation capabilities while navigation. 
Moving forward, it is imperative for future research to integrate foundation models for situated task-oriented dialogue management~\citep{ulmer2024bootstrapping}, or explore existing foundation models for task-oriented dialogue~\citep{he2022galaxy}.

\paragraph{Agent Model: Adapting Foundation Models for VLN.} 
While foundation models show strong generalizability, incorporating them into navigation tasks remains challenging. 
LLMs fundamentally lack the capability to visually perceive the actual environment and are prone to hallucinations. We als discuss capabilities of LLMs in planning and reasoning.
\vspace*{-5pt}
\begin{itemize}[leftmargin=*]
\setlength\itemsep{-0.25em}
\item \textit{Lack of Embodied Experience.} 
This limitation can lead to scenarios where LLMs rely solely on pre-established commonsense for task planning and reasoning, which might not meet specific real-world needs~\citep{xiang2024language}.
Some pipelines tackle this issue by captioning the visual observations to textual descriptions as prompts for LLMs~\citep{zheng2022jarvis}, with a potential loss of essential visual semantics.
Compared with LLMs, VLM agents demonstrate the potential to perceive the visual world and plan~\citep{zhang2024navid}.
Still, these models are primarily developed from internet data, which lack embodied experiences~\citep{mu2024embodiedgpt} and need finetuning for robust agentic decision-making~\citep{zhai2024fine}.
Further research is needed to transfer the commonsense knowledge in foundation model agents to generalize to embodied situations. 
Recently proposed embodied foundation models~(such as EmbodieGPT~\citep{mu2024embodiedgpt}, PaLM-E~\citep{driess2023palm} and Octopus~\citep{yang2025octopus}) offer a promising solution for enabling agents to operate more effectively in interactive environments. They fine-tune foundation models across multiple embodied tasks to bridge the gap between an agent's understanding of vision, language, and embodied actions, enhancing the foundation model's ability to comprehend and execute based on multimodal input.

\item \textit{Hallucination Issue.}
LLMs and VLMs might generate non-existent objects, leading to misinformation~\citep{li2023evaluating,chen2024multiobject}.
For example, when LLM performs task planning, it may generate instructions such as ``\textit{go forward and turn left at the sofa}'' even if there is no sofa in the room. This inaccuracy may cause them to execute incorrect or impossible actions.

\item \textit{LLMs in Planning and Reasoning.}
There are some literatures evaluating the zero-shot reasoning and planning capabilities of LLMs, particularly in relation to PlanBench~\citep{Valmeekam2022PlanBenchAE} and CogEval~\citep{Momennejad2023EvaluatingCM}, which highlight LLMs’ limitations in more complex planning tasks. These works assess LLMs in a variety of challenging settings, such as plan generation, optimality, robustness, and reasoning, and identify that LLMs sometimes struggle with hallucinations or fail to grasp the relational structures underlying complex planning problems.
In the context of VLN, the action space and the planning requirements are relatively constrained due to the fixed indoor environments and the limited set of navigational actions. This bounded setting makes it more feasible for LLMs to provide step-by-step instructions for coarse-grained directions, which has been demonstrated to be effective in previous works. In VLN tasks, the LLM’s role is not to take over the entire planning process, but rather to assist by offering a structured breakdown of instructions. The agent’s actual decision-making remains primarily reliant on other components, such as perception and motion control. Therefore, in VLN tasks, the LLM’s planning serves more as a supplementary guide rather than the sole decision-making factor.
\end{itemize}

\paragraph{Deployment: From Simulation to Real Robots.} 
Simulated settings often lack the complexity and variability of real-world environments, and lower-quality rendered images exacerbate this issue. 
First, the perception gap results in decreased performance and accuracy, highlighting the need for more robust perception systems. 
\citet{wang2024sim} have started to explore the use of semantic maps and 3D feature fields to provide monocular robots with panoramic perception shows improved performance.
The embodiment gap and the data scarcity are also bottlenecks.
The rise of robot teleportation~\citep{he2024learning} also provides an alternative to scale up VLN data for foundation models in real human-robot communications.

\section{Broader Impact}
Foundation models hold great promise for advancing vision-language navigation. However, it is essential to address their broader ethical, legal, and societal implications. 
Given that they are pre-trained on vast, web-scale datasets, these models can carry inherent biases, which may result in fairness concerns, \textit{e.g.}, to multilingual users.
Some approaches involve continual model training, it is critical to acknowledge and mitigate any potential risks to user privacy, especially when deployed in real-world applications such as home robotics.
\section{Acknowledgement}
This work is supported in part by the ARO Award W911NF2110220, NSF grant IIS-1949634, and ONR grant N00014-23-1-2417 \& N00014-23-1-2356.
Any opinions, findings, and conclusions or recommendations expressed in this material are those of the authors and do not necessarily reflect the views of the funding agencies.

\bibliography{main}
\bibliographystyle{tmlr}

% \appendix
% \section{Appendix}
% You may include other additional sections here.

\end{document}